\documentclass[letterpaper, 10 pt, conference]{ieeeconf}  

\usepackage{cite}
\usepackage{amsmath,amssymb,amsfonts}
\usepackage{algorithmic}
\usepackage{graphicx}
\usepackage{textcomp}
\usepackage{xcolor}
\usepackage{color,soul}
\usepackage{svg}

\IEEEoverridecommandlockouts                              

\title{\LARGE \bf Attention-enabled Explainable AI for \\Bladder Cancer Recurrence Prediction}

\author{Saram Abbas$^{1,*}$, Naeem Soomro$^{2}$, Rishad Shafik$^{1}$, Rakesh Heer$^{3,4}$, Kabita Adhikari$^{1,*}$
\thanks{This work was not supported by any organization}
\thanks{$^{1}$Saram Abbas, Rishad Shafik and Kabita Adhikari are with the School of Engineering, Newcastle University, Newcastle upon Tyne, UK. \tt\small s.abbas11@newcastle.ac.uk, rishad.shafik@newcastle.ac.uk, kabita.adhikari@newcastle.ac.uk}%
\thanks{$^{2}$Naeem Soomro is with the Department of Urology, Freeman Hospital, Newcastle upon Tyne, UK. \tt\small n.soomro@nhs.net}%
\thanks{$^{3}$Rakesh Heer is with the Division of Surgery, Imperial College London, London, UK. \tt\small r.heer@imperial.ac.uk}%
}

\begin{document}

\maketitle
\thispagestyle{empty}
\pagestyle{empty}

\begin{abstract}
        Non-muscle-invasive bladder cancer (NMIBC) is a relentless challenge in oncology, with recurrence rates soaring as high as 70–80\%. Each recurrence triggers a cascade of invasive procedures, lifelong surveillance, and escalating healthcare costs—affecting 460,000 individuals worldwide. However, existing clinical prediction tools remain fundamentally flawed, often overestimating recurrence risk and failing to provide personalized insights for patient management. In this work, we propose an interpretable deep learning framework that integrates vector embeddings and attention mechanisms to improve NMIBC recurrence prediction performance. We incorporate vector embeddings for categorical variables such as smoking status and intravesical treatments, allowing the model to capture complex relationships between patient attributes and recurrence risk. These embeddings provide a richer representation of the data, enabling improved feature interactions and enhancing prediction performance. Our approach not only enhances performance but also provides clinicians with patient-specific insights by highlighting the most influential features contributing to recurrence risk for each patient. Our model achieves accuracy of ~70\% with tabular data, outperforming conventional statistical methods while providing clinician-friendly patient-level explanations through feature attention.  Unlike previous studies, our approach identifies new important factors influencing recurrence, such as surgical duration and hospital stay, which had not been considered in existing NMIBC prediction models.
\end{abstract}

\section{INTRODUCTION}

Non-muscle-invasive bladder cancer (NMIBC) ranks as the ninth most common malignancy globally, posing an enduring challenge to healthcare systems due to its high recurrence rates \cite{BladderCancerStatistics2015}. Patients diagnosed with NMIBC require lifelong surveillance, which, if not accurately risk-stratified, can lead to excessive costs, overtreatment or even progress to a much deadlier form -- muscle-invasive bladder cancer \cite{vandenboschLongtermCancerspecificSurvival2011,hallGuidelineManagementNonmuscle2007}. Bladder cancer even cost the EU up to €4.9 billion in 2012 \cite{lealEconomicBurdenBladder2016,mossanenBurdenBladderCancer2014}. Current standard predictive tools, such as EORTC or CUETO, often yield concordance indices of only 0.55–0.61, indicating poor discriminatory power \cite{vedderRiskPredictionScores2014}. The concordance index ranges from 0.5 (no better than random chance) to 1.0 (perfect prediction), meaning these tools perform only marginally better than guessing. These have poor accuracy due to failing to incorporate temporal disease progression, longitudinal treatment effects, and complex interactions between clinical and molecular factors, thereby limiting their predictive reliability and clinical utility \cite{utsumiDevelopmentValidationClinical}.

While machine learning (ML) and deep learning have shown potential to surpass traditional methods in adaptability and predictive accuracy, they too exhibit notable limitations. A significant drawback of most ML approaches is their inability to provide insights into the importance of individual features, as they often lack mechanisms, such as attention, to explicitly prioritize or explain the impact of key clinical variables. Without attention mechanisms, features are processed uniformly during input, and while models can still implicitly learn feature importance, it is challenging to directly interpret which feature or combination of features contributes most to the predictions. This lack of transparency reduces trust among clinicians, who demand models that not only perform well but also offer clear, interpretable explanations.

Furthermore, conventional ML models currently used in NMIBC recurrence prediction often treat categorical data, such as smoking status, tumour grade, or intravesical treatments, with basic encodings like one-hot encoding. While effective for representing categorical variables, one-hot encoding fails to capture nuanced relationships between categories, such as the relative effectiveness of treatment regimens or the varying risks associated with tumour grades. Although neural networks can learn such associations over time, one-hot encoding offers no inherent structure to support this learning. In contrast, vector embeddings—an established technique widely adopted in domains like natural language processing—enable the model to learn a continuous representation of categorical variables, capturing nuanced similarities and hierarchical relationships. This study adopts embeddings as a means of better encoding for structured domain knowledge and improve recurrence prediction performance in a clinical context that has seen limited application of such methods.

In this work, we break new ground by designing a deep learning framework that uniquely integrates attention mechanisms and vector embeddings to tackle the specific challenges of NMIBC recurrence prediction. Our model not only uncovers hidden patterns in heterogeneous clinical data—spanning intravesical treatments, tumour profiles, and lifestyle factors—but also delivers patient-specific insights that mirror the nuanced decision-making of experienced oncologists. By prioritising features dynamically through attention, the model allows clinicians to understand exactly why a prediction is made, bridging the gap between performance and transparency. This marks a significant step towards transforming NMIBC management with AI that clinicians can trust, offering both precision and interpretability in a way traditional models cannot achieve.

The remainder of this paper is structured as follows. Section II reviews related work in deep learning for tabular clinical data and NMIBC recurrence prediction. Section III describes the dataset, preprocessing strategies, and the proposed model architecture. Section IV presents the results, including predictive performance and feature importance analysis. Section V discusses the broader significance, limitations, and potential future directions. Finally, Section VI serves as the conclusion for the paper.

\section{RELATED WORK}
        This work sits at the intersection of oncology, structured clinical data, and deep learning. In the following, we review advances in deep learning for tabular data, and discuss prior approaches to recurrence prediction in NMIBC to position our contribution within the existing landscape.
        \subsection{Tabular Data in Medical AI}

                Despite being the foundation of clinical decision-making, tabular data has often been overshadowed by deep learning’s successes in imaging and unstructured text analysis. Patient demographics, biomarkers, and surgical outcomes are structured, yet neural networks struggle with them due to the lack of spatial or sequential patterns that CNNs and transformers rely on \cite{borisovDeepNeuralNetworks2024, shwartz-zivTabularDataDeep2022}. This challenge has kept traditional models—logistic regression, random forests, and gradient boosting—as the standard in clinical predictions. While effective, these methods depend on extensive feature engineering and fail to adapt dynamically to complex, multi-modal patient data \cite{renDeepLearningTabular2025}.

        \subsection{Advances in Deep Learning for Tabular Data}

                Deep learning’s response to this challenge has been the development of architectures designed specifically for structured data. Attention-based models like SAINT \cite{somepalliSAINTImprovedNeural2021} and TabNet \cite{arikTabNetAttentiveInterpretable2021} allow models to focus on the most relevant features dynamically. Transformer-based models such as TabTransformer \cite{huangTabTransformerTabularData2020} and GaBERT \cite{houGaBERTInterpretablePretrained2024} further improve representation learning, capturing complex feature interactions in ways that traditional tree-based models cannot. However, while these models show promise, their adoption in structured clinical data remains slow, and they often struggle to consistently outperform traditional approaches in real-world applications \cite{chengMoAGLSAMultiomicsAdaptive2024}.

        \subsection{Applications in Oncology and NMIBC}

                In oncology, deep learning has demonstrated success in breast and lung cancer prognosis, yet its use in NMIBC remains limited \cite{jungCancerGATEPredictionCancerdriver2024a, lanDeepKEGGMultiomicsData2024}. Current NMIBC recurrence prediction relies on conventional risk scores (e.g., EORTC) and machine learning models, which, while clinically useful, fail to fully capture the complex, nonlinear relationships between tumor characteristics, treatments, and patient history. 
                
                From the most recent systematic reviews \cite{abbasAIPredictingRecurrence2025, borhaniArtificialIntelligencePromising2022, kluthPrognosticPredictionTools2015}, only four studies were found to rely solely on clinical, treatment, or demographic variables for NMIBC recurrence prediction using machine learning. Ajili et al. \cite{ajiliPrognosticValueArtificial2014} employed a multilayer perceptron trained on post-BCG immunotherapy patient data, achieving high sensitivity (96.7\%) and specificity (100\%), but their findings were based on a very small dataset (n=40), severely limiting generalisability. Zhang and Ma used \cite{zhangPredictiveValueTotal2024} conventional classifiers to assess recurrence risk using serum biomarkers (CA50 and total bilirubin), reporting only modest AUC values (up to 0.623), indicating insufficient discriminative power for clinical use. Schwarz et al. \cite{schwarzRelevantFeaturesRecurrence2024} applied SVM, ANN, and gradient boosting to clinical variables and reported strong performance (F1-score 83.9\%, AUC 70.8\%). While they attempted to improve interpretability using permutation feature importance, their model still lacked patient-specific reasoning, making it difficult to integrate into clinical workflows. Lastly, Jobczyk et al. \cite{jobczykDeepLearningbasedRecalibration2022} developed a DeepSurv model recalibrating EORTC/CUETO risk scores using one of the largest NMIBC datasets to date (n=3,892). Their recurrence prediction performance (C-index ~0.65) improved upon traditional tools, but still left considerable room for improvement, particularly in model explainability.

                These studies show that while ML can outperform traditional tools using clinical data alone, they fall short on key fronts: they don’t prioritize patient-specific features, lack interpretability, and treat variables as isolated, missing the complex, real-world interactions that shape recurrence risk.

\begin{figure}

        \centering
        \includegraphics[width=0.45\textwidth, height=15cm]{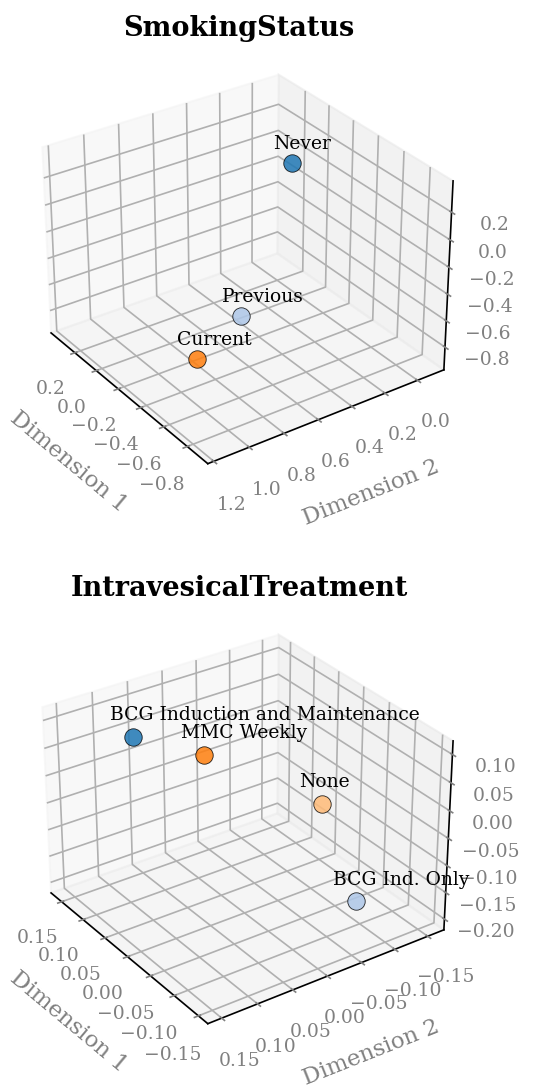}

        \caption{3D visualization of vector embeddings for categorical features, showing recurrence risk patterns learned by the AI model. In the IntravesicalTreatment plot (bottom), "BCG Induction Only" is farthest from the origin, indicating the highest recurrence risk. "None" is at the origin, suggesting it as the baseline risk, while "BCG Induction and Maintenance" shows the lowest recurrence risk. In the SmokingStatus plot (top), "Never" smokers are farthest from the other categories, indicating the lowest risk, while "Current" smokers are closest to the origin, suggesting the highest risk. "Previous" smokers lie in between, reflecting an intermediate risk level.}
        \label{fig:vector_embeddings}
\end{figure}

\section{METHODS}
        In this section, we describe the dataset, preprocessing steps, and the design of our deep learning model. Particular focus is given to how vector embeddings and attention mechanisms are integrated to enhance prediction performance and interpretability. 

        \subsection{Data Description and Preprocessing}
                Our dataset stands out in NMIBC research for its diversity and depth, offering a breadth rarely seen in similar studies. Unlike conventional datasets that narrowly focus on tumour characteristics, ours is enriched with a wide range of clinical, demographic, and lifestyle features, creating a comprehensive view of each patient. This includes tumour grade, stage, and tumour number as well as lifestyle factors such as smoking history and quality-of-life metrics (EQ-5D), alongside procedural details like surgical time and total days in hospital—features that are often overlooked but crucial to understanding recurrence risk.

                The data for this study was derived from a subset of patients enrolled in the PHOTO RCT \cite{rakeshPhotodynamicWhitelightguidedResection}, a multicentre, pragmatic, open-label randomized controlled trial conducted across 22 NHS hospitals in the UK. Over 500 patients diagnosed with NMIBC were recruited and monitored for up to 36 months. Patients underwent regular check-ups every 3 to 6 months, with records maintained on tumour recurrence, progression, and treatment outcomes.

                Our data combines structured clinical data with variables reflecting lifestyle and procedural factors, enabling the discovery of previously hidden patterns. For instance, our inclusion of procedural factors like surgical duration provides a unique perspective on how treatment logistics impact recurrence—a perspective largely absent in existing models.

                From the original cohort of 500 NMIBC patients, a total of 296 patients were retained. Exclusions were due to missing or incomplete data, progression to muscle-invasive disease, and patient dropout during follow-up. Alternative imputation techniques were evaluated; however, they did not yield satisfactory performance or stability. Therefore, listwise deletion was deemed the most appropriate approach. The resulting dataset comprized 40\% recurrence cases and 60\% non-recurrence cases. To address this slight class imbalance, we employed the Synthetic Minority Over-Sampling Technique (SMOTE) which generates synthetic examples of the minority class by interpolating between existing minority instances and their nearest neighbors in the feature space. This ensured better representation of recurrence cases during model training. This resulted in a final dataset with 356 patients after SMOTE.

                The final dataset included 23 variables: 5 numerical, 6 categorical, and 12 binary. Each type underwent tailored preprocessing. Numerical variables (e.g., age, surgical time) were scaled, normalized, and subjected to outlier detection and removal to ensure stable gradient flow during training. Categorical variables (e.g., smoking status, tumour grade) were transformed using vector embeddings, enabling the model to capture semantic relationships between categories. Binary features were retained in their native binary form, as further transformation was unnecessary.

        \subsection{Neural Network Architecture with Vector Embeddings}
                To effectively represent heterogeneous data, we utilized vector embeddings for categorical features. Unlike one-hot encoding (the most common encoding)—which assigns a discrete vector with no notion of proximity or severity differences—embeddings map categories into a hyper-dimensional vector space, {capturing subtle hierarchies and interlinked relationships critical for clinically relevant predictions.} For instance, as shown in Fig.~\ref{fig:vector_embeddings}, the model captures how recurrence risk does not change uniformly across smoking categories—for example, the reduction in risk when moving from a smoker to a former smoker is more pronounced than the increase in risk when moving from a former-smoker to a current-smoker. For each categorical variable, we created a trainable embedding matrix that maps each category to a hyper-dimensional vector. Embeddings were learned end-to-end during training via backpropagation. This richer representation enables the network to identify nuanced clinical patterns in the data, ultimately improving predictive accuracy.

\begin{figure}[t]
                \centering
                \includegraphics[width=0.45\textwidth]{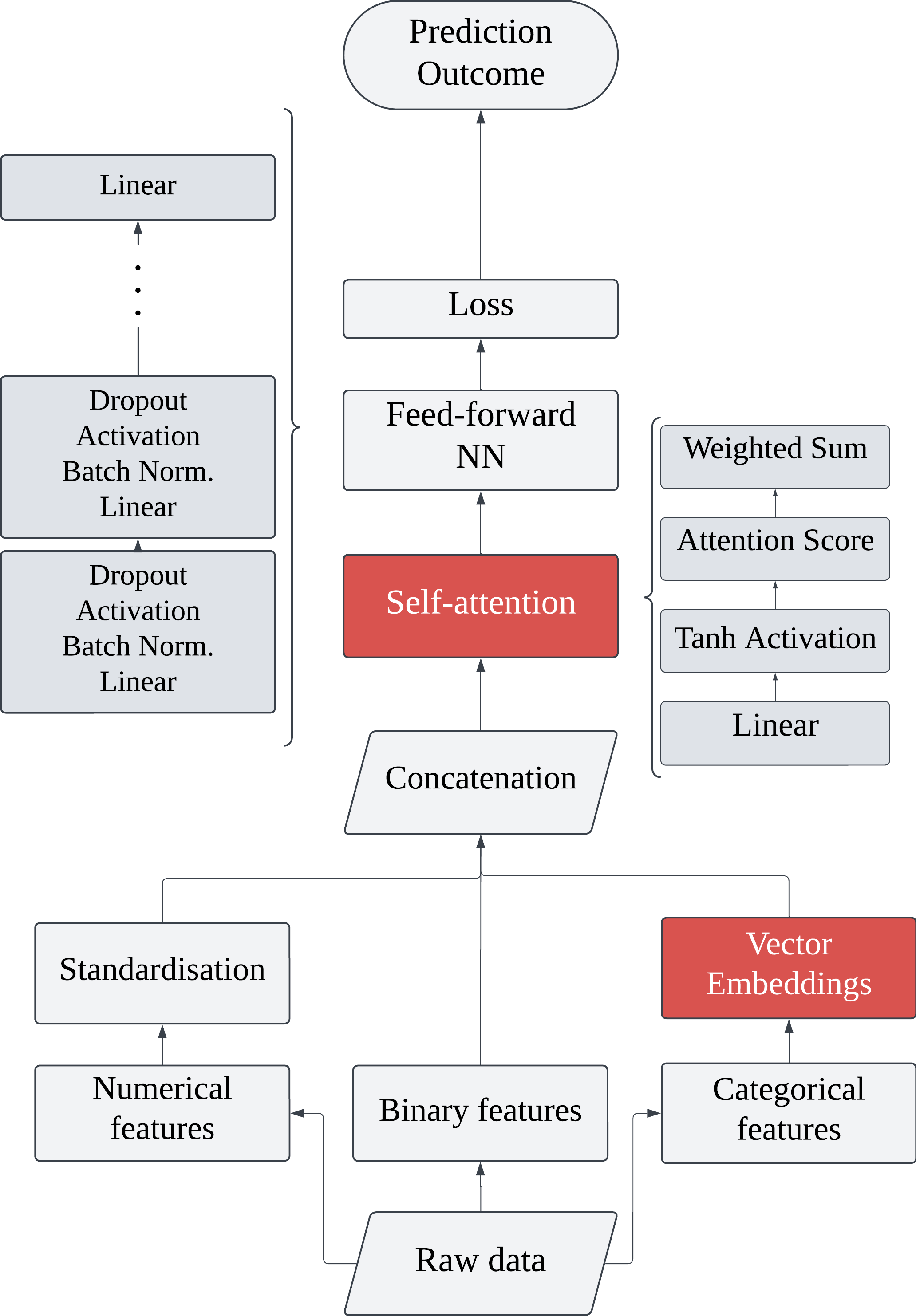}
                \caption{Neural Network Architecture with Vector Embeddings and Self-Attention used in this work}
                \label{fig:nn_architecture}
                \end{figure}

                The proposed network architecture combines feature embeddings and an attention mechanism to enhance predictive accuracy and interpretability in bladder cancer recurrence prediction. Input data—including categorical, numerical, and binary features—is processed into dense vector embeddings, ensuring compatibility and effective representation of heterogeneous data types. These embeddings are concatenated into a unified feature vector and passed through an attention mechanism, which assigns importance weights to individual features, highlighting clinically relevant predictors. The weighted representation is fed into dense layers with ReLU activations, dropout regularization, and batch normalization to improve generalization. The output layer utilizes a sigmoid activation to predict recurrence probability. This architecture uniquely integrates {vector embeddings} with attention mechanisms to offer both high performance and transparency, providing valuable insights into feature contributions for clinical decision-making. The network architecture is visualized in Fig.~\ref{fig:nn_architecture}.

                Several alternative architectures were also explored, including standard fully connected networks without attention and transformer-based models. To ensure optimal performance, we performed model architecture tuning using the Optuna framework \cite{akibaOptunaNextgenerationHyperparameter2019}, which allowed for automated hyperparameter optimisation across a defined search space. The final model architecture—embedding-based with feature-level attention—was selected based on superior validation performance and its ability to provide interpretable, patient-specific predictions.

\subsection{Attention Mechanisms}
        \begin{figure*}[bthp]
                \centering
                \includegraphics[width=0.9\textwidth]{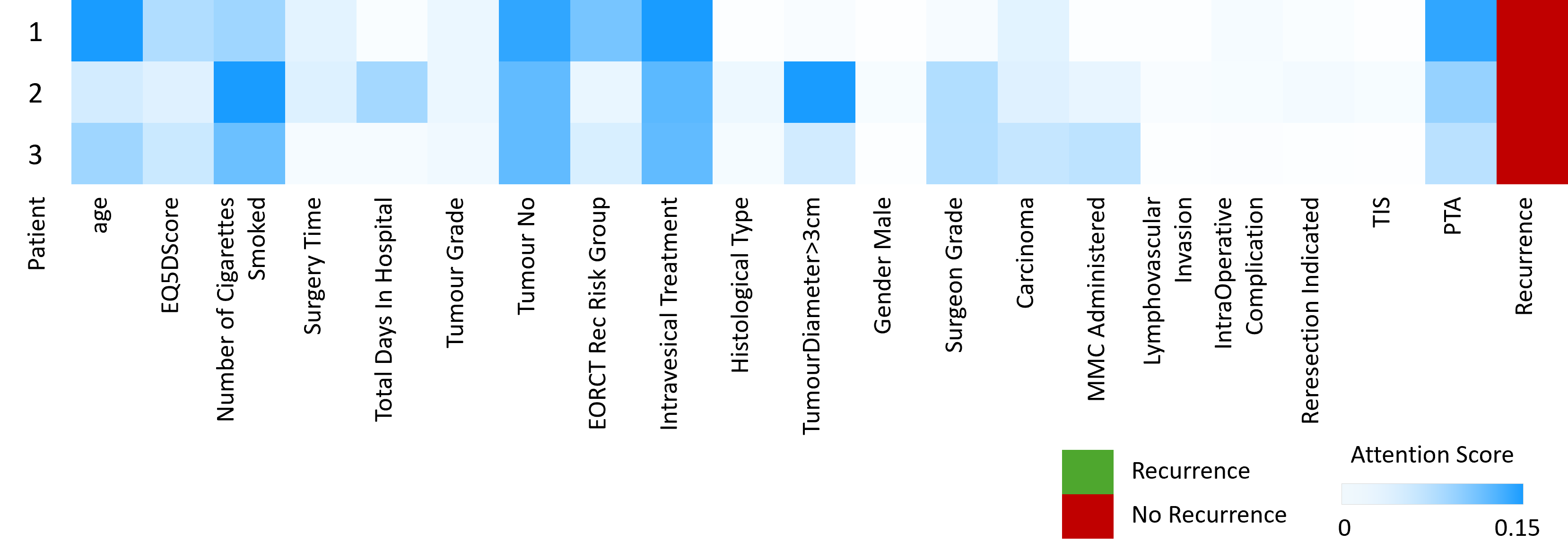}%
                \caption{Snippet of the patient-level attention heatmap, where each row represents a patient and each column corresponds to a unique clinical or demographic variable. Darker cells represent stronger attention weights, indicating where the model "looks" more intently for each individual’s recurrence risk. The red box in the Recurrence column designates that these patients are predicted to experience recurrence. Notably, although they share the same outcome, each patient’s attention weights differ across various features, highlighting the individualized nature of the model’s analysis. This transparent focus mimics an experienced clinician's reasoning to pinpoint the most influential predictors for any given patient.}
                \label{fig:heatmap}
        \end{figure*}

        Recent studies on tabular data have explored attention-based architectures to improve both performance and interpretability \cite{somepalliSAINTImprovedNeural2021,arikTabNetAttentiveInterpretable2021}. In our work, we build on this concept by introducing a \emph{feature-level attention module} in the pipeline of the network to dynamically prioritize features in the context of NMIBC recurrence prediction. While SAINT and TabNet have demonstrated promising results in various healthcare applications, to our knowledge, this is the first effort to apply a similar attention mechanism specifically for NMIBC recurrence risk estimation.

        To formalize the approach, we define
        \[
        X = [x_1, x_2, \dots, x_n]
        \]
        as the concatenated embeddings of $n$ features, where each embedding $x_i \in \mathbb{R}^d$ is a $d$-dimensional vector capturing clinical and demographic information. We compute an alignment score $s_i$ for each feature embedding:
        \[
        s_i = \mathbf{w}^\mathsf{T} \tanh\!\bigl(\mathbf{W} x_i + \mathbf{b}\bigr),
        \]
        where $\mathbf{W} \in \mathbb{R}^{k \times d}$, $\mathbf{b} \in \mathbb{R}^{k}$, and $\mathbf{w} \in \mathbb{R}^k$ are trainable parameters, and $k$ controls the dimensionality of the intermediate attention space. The scalar $s_i$ reflects the preliminary importance of feature $i$ relative to the others.

        Next, we apply the softmax function to normalize the alignment scores into attention weights $\alpha_i$:
        \[
        \alpha_i = \frac{\exp(s_i)}{\sum_{j=1}^{n} \exp(s_j)}, 
        \quad \text{where} \quad \sum_{i=1}^{n} \alpha_i = 1.
        \]
        The softmax ensures that $\alpha_i \in [0,1]$ and collectively sum to 1, thus allowing them to be interpreted as probabilities or relative importance values for each feature embedding $x_i$.

        Finally, we obtain the \emph{attention-weighted representation} $h \in \mathbb{R}^d$ as:
        \[
        h = \sum_{i=1}^{n} \alpha_i \, x_i.
        \]
        By aggregating features in this manner, the model can focus on those most pertinent to the clinical outcome under consideration. For example, a patient with extensive smoking history may elicit larger $\alpha_i$ values for smoking-related variables, emphasizing their well-known influence on NMIBC recurrence risk.

        To provide insight into the model's reasoning, we visualized the learned attention patterns as heatmaps at the individual patient level. These visualizations provide a direct indication of how much weight the network assigns to each clinical or demographic feature. For example see Fig.~\ref{fig:heatmap}, in patient 1, the model focuses on age, tumour number, intravesical treatment, and PTA as key predictors, whereas for patient 2, it prioritizes the number of cigarettes smoked and tumour diameter. Although both patients experienced recurrence, the model evaluates them differently based on their unique clinical profiles. {This clearly sets our model apart by showing its ability to personalize predictions for each patient—something traditional machine learning models simply cannot do. Unlike the generic, one-size-fits-all methods used today, our approach adapts to the nuances of individual patient data.}

\section{RESULTS}
        In this section, we evaluate the model’s predictive performance, analyze feature importance through attention weights, and demonstrate how the model provides patient-specific explanations.
        \subsection{Model Training Performance}

\begin{figure}[tbp]
        \centering
        \includegraphics[width=0.5\textwidth]{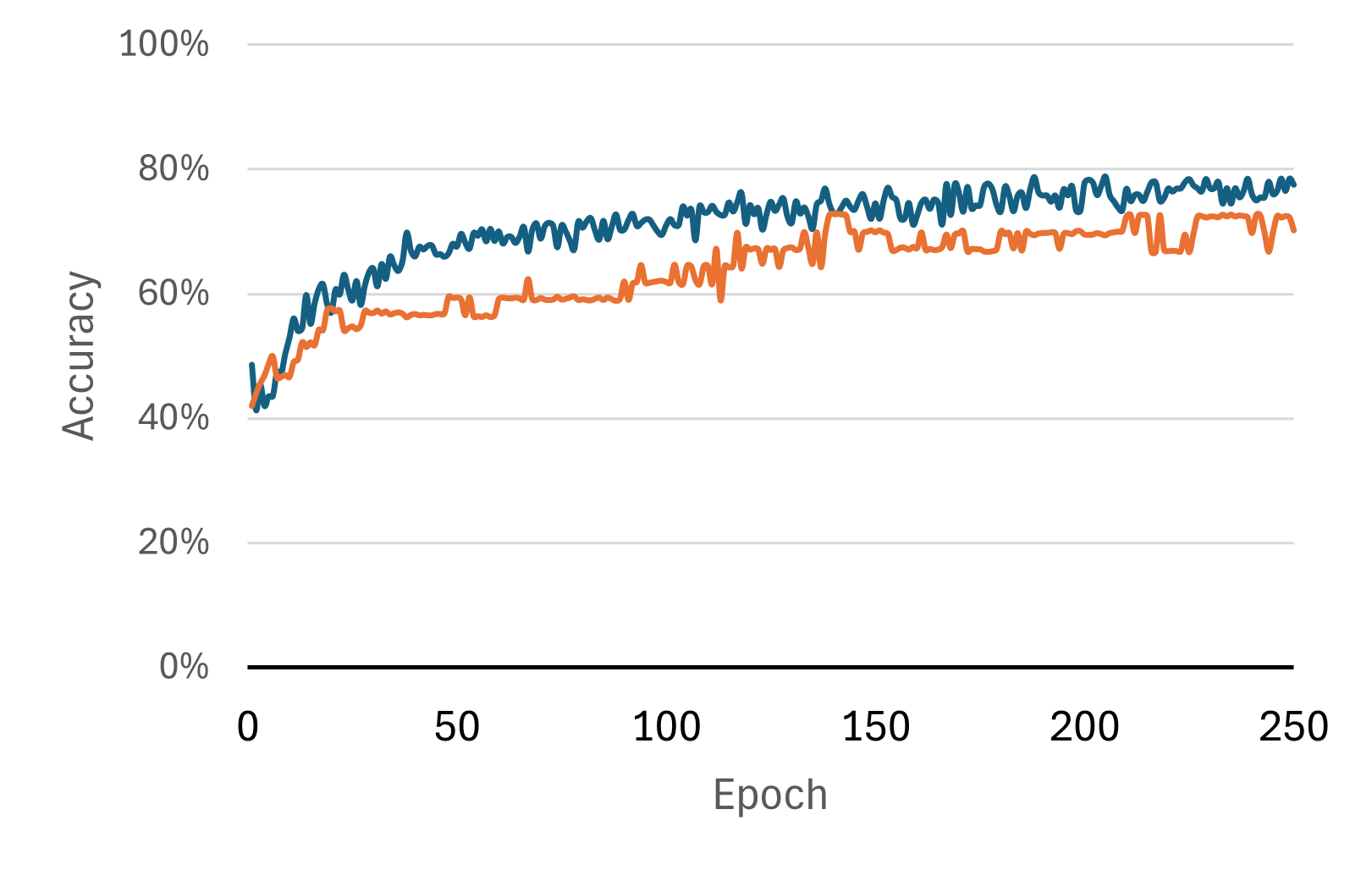}%
        \caption{Training (blue) and validation (orange) accuracy across 250~epochs for the proposed neural network. The model achieves a final validation accuracy of about 70\%, with training accuracy levelling at around 80\%.}
        \label{fig:accuracy}
        \end{figure}

                The model’s training and validation accuracy were tracked over 250 epochs, shown in Fig.~\ref{fig:accuracy}. Initially, both curves rose sharply (from about 20--30\%), suggesting the model was quickly learning basic patterns. After around 50 epochs, validation accuracy began to stabilize between 50--60\%. Eventually, both curves converged near epoch 200, culminating in a validation accuracy plateau of roughly 70\%. The gap between training and validation did not widen drastically, implying that the network generalizes reasonably well despite the relatively small dataset of 356 patients. This architecture showed markedly improved performance compared to other algorithms such as Logistic regression (52\%), simple feedforward NN without attention or embeddings (55\%) and TabNet (63\%).  

        \begin{figure}[thb!]
        \centering
        \includegraphics[width=0.5\textwidth]{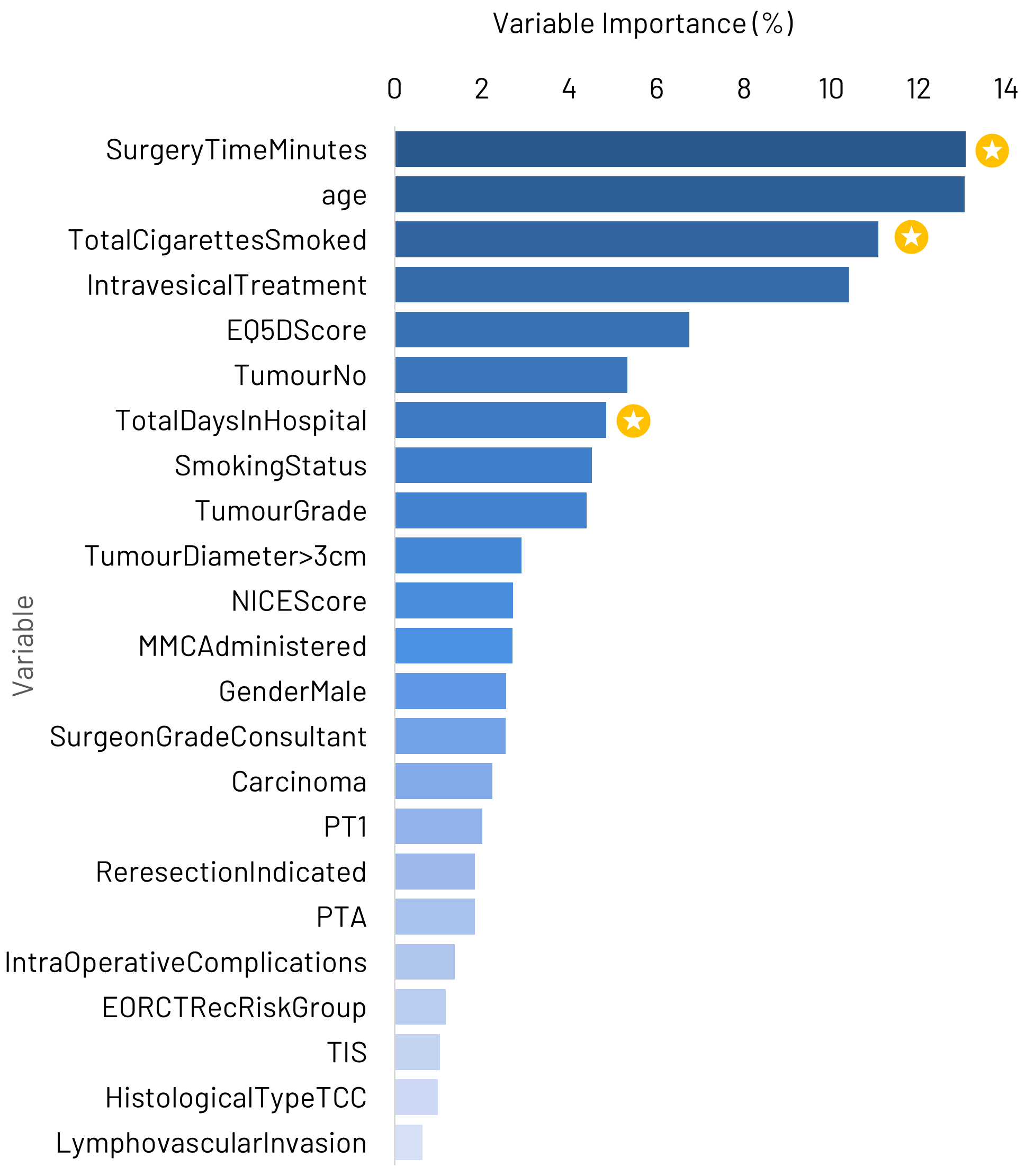}
        \caption{Feature importance rankings from the attention-based neural network for NMIBC recurrence prediction. The gold stars indicate newly recognized variables—such as Surgery Time, Total Number of Cigarettes Smoked, and Total Days in Hospital—which were not prioritized in earlier NMIBC studies. This novel ordering highlights the capacity of modern attention mechanisms to uncover clinically significant factors beyond those identified by conventional statistical or machine learning models.}
        \label{fig:importance}
        \end{figure}

        \subsection{Feature Importance Analysis}

                Fig.~\ref{fig:importance} presents the ranked importance of features derived from attention weights. Notably, \emph{surgical time}, \emph{age}, and \emph{total cigarettes smoked} emerge as the three most influential factors. This outcome aligns partially with existing clinical wisdom—older age and smoking history are established risk indicators—while also highlighting procedural aspects (e.g., longer surgeries) that traditional scoring systems often overlook. Intravesical therapy, tumour count, and overall health metrics (e.g., EQ5DScore) were also among the top contributors, reinforcing the notion that recurrence risk is multifactorial. 

        \subsection{Attention-Based Interpretability}

                To complement aggregate feature rankings, we also examined patient-level attention distributions. As shown in Fig.~\ref{fig:heatmap}, each row represents a single patient, and each column corresponds to a unique feature. Darker cells denote higher attention weights, implying stronger influence on the model’s final prediction. For instance, if a patient had a significant smoking history and a high-grade tumour, those factors receive more attention (darker blue), overshadowing other attributes like gender or tumour diameter for that particular case.

                Such visualized weighting helps clinicians trace how certain patient profiles—e.g., heavy smokers with longer operative times—can drastically alter risk. By elucidating which variables “matter” most for each individual, the model addresses interpretability concerns, potentially leading to more patient-tailored surveillance protocols and improved clinical trust.

        \subsection{Discussion of key Findings}

                These results demonstrate that the proposed attention-enabled embedding model not only achieves a 70\% validation accuracy in NMIBC recurrence prediction but also provides clinician-friendly explanations. Key factors like surgery duration and smoking status—which are traditionally underrepresented in static scoring tools—emerge as notable contributors to recurrence risk. Equally important, the attention heatmap offers a transparent view into the model’s reasoning, facilitating better acceptance and potential adoption in clinical workflows.

\section{DISCUSSIONS}
        Here we discuss how our findings advance NMIBC recurrence prediction, explore their broader clinical implications, address current limitations, and propose directions for future research to build on this foundation.
        
        \subsection{Significance}
                Our framework aligns with the urgent clinical need for improved, interpretable NMIBC recurrence prediction. By marrying advanced ML techniques with clinician-centric explanations, we offer a tool capable of guiding surveillance intervals, informing intravesical therapy decisions, and optimising resource allocation. Minimising overtreatment and improving patient quality of life serve as key downstream benefits of more accurate and trustworthy risk stratification.

        \subsection{Broader Implications and Limitations}
                Though focused on NMIBC, these methods generalize to other cancers and diseases marked by high recurrence rates. Integrating longitudinal data and refining imputation methods may further enhance predictive fidelity. Collaboration among institutions would enable much larger datasets hence increasing statistical power and enabling the model to transcend current performance ceilings.

                Challenges remain in this study. The modest dataset size and missing data restricted accuracy. Future work includes employing more sophisticated data augmentation, exploring more advanced attention variants, and integrating external clinical knowledge bases to inform the embedding space.

        \subsection{Future Work}
                Expanding datasets, incorporating genomic or imaging biomarkers, and utilising explainability techniques like SHAP or LIME can complement attention-driven interpretations. In this work, we chose to focus on attention-based feature importance, as it is integrated directly into the model and allows both global and patient-specific explanations during inference. This avoids the need for separate post hoc analysis and offers a more seamless way to interpret predictions, especially for clinicians. However, we acknowledge that SHAP remains a valuable and widely-used tool, and we plan to explore it in future work to validate and compare against the attention-based rankings we report here. In future work, we also aim to incorporate inherently interpretable models, such as the Tsetlin Machine, which leverages propositional logic to provide transparent, rule-based explanations without needing attention mechanism or vector embeddings.

\addtolength{\textheight}{-9cm}   


\section{CONCLUSION}
        We have demonstrated that combining vector embeddings and attention mechanisms can significantly enhance NMIBC recurrence prediction accuracy while retaining interpretability. By integrating vector embeddings and attention mechanisms, our model achieves a validation accuracy of approximately 70\%, outperforming current statistical methods being used in practice such as EORTC and CUETO. A key achievement of this work is the identification of novel predictors, such as surgical duration and hospital stay, which were previously overlooked but have now been shown to influence recurrence risk. Our findings also demonstrate that the model can assess patients individually—considering different clinical and demographic factors for each case—mimicking the nuanced decision-making process of experienced oncologists.

        Beyond predictive performance, this work achieves a crucial breakthrough in AI explainability for bladder cancer. For the first time, clinicians can pinpoint exactly which features contributed to a prediction, transforming AI from an opaque "black box" into a transparent and actionable tool. This transparency fosters trust, enabling more personalized surveillance strategies and targeted treatment planning.

\bibliographystyle{IEEEtran}
\bibliography{Attention_Conference_paper}
\end{document}